# Variational Inference in Non-negative Factorial Hidden Markov Models for Efficient Audio Source Separation

**Gautham J. Mysore**  GMYSORE@ADOBE.COM
Advanced Technology Labs, Adobe Systems Inc., San Francisco, CA 94103, USA

**Maneesh Sahani**  MANEESH@GATSBY.UCL.AC.UK
Gatsby Computational Neuroscience Unit, University College London, WC1N 3AR, UK

## Abstract

The past decade has seen substantial work on the use of non-negative matrix factorization and its probabilistic counterparts for audio source separation. Although able to capture audio spectral structure well, these models neglect the non-stationarity and temporal dynamics that are important properties of audio. The recently proposed non-negative factorial hidden Markov model (N-FHMM) introduces a temporal dimension and improves source separation performance. However, the factorial nature of this model makes the complexity of inference exponential in the number of sound sources. Here, we present a Bayesian variant of the N-FHMM suited to an efficient variational inference algorithm, whose complexity is linear in the number of sound sources. Our algorithm performs comparably to exact inference in the original N-FHMM but is significantly faster. In typical configurations of the N-FHMM, our method achieves around a 30x increase in speed.

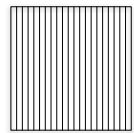

Figure 1. Conceptual model of a single sound source using non-negative matrix factorization. A single dictionary is used to explain the entire sound source.

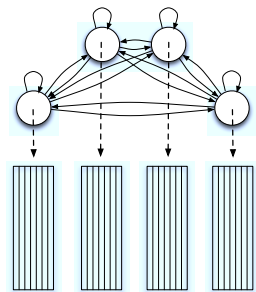

Figure 2. Conceptual model of a single sound source using the non-negative hidden Markov model. Multiple dictionaries account for non-stationarity and the Markov chain accounts for temporal dynamics.

## 1. Introduction

Spectrograms reveal a great deal of acoustic structure and are therefore often the representation of choice for modeling sounds. A spectrogram is the magnitude of the short-time Fourier transform (STFT) of a signal and is therefore a non-negative matrix. This has led to the popularity of using non-negative matrix factorization (NMF) (Lee & Seung, 2001) to model audio (Smaragdis & Brown, 2003).

Conceptually, NMF models each time frame (column) of an audio spectrogram as a linear combination of non-negative dictionary elements. Given the spectrogram of a sound source, we can use NMF to learn a dictionary (Figure 1) that serves as a model for the range of different short-term spectra generated by that source. While potentially a powerful spectral model, NMF provides no generative account of the temporal dynamics linking these short-term spectra, nor of any spectral non-stationaries: two essential ingredients of real audio signals.

A recent proposal, the non-negative hidden Markov model (N-HMM) (Mysore et al., 2010), deals with this failing of NMF by modeling a sound source with multiple dictionaries (Figure 2) such that each time frame of the spectrogram is modeled by a linear combination of the elements of any one of its many dictionaries, essentially allowing different sub-models for spectra at different time frames. Moreover, transitions between these component dictionaries from one time frame to the next are governed by a Markov chain, thus capturing the temporal dynamics of the source. The model is described more completely in Section 2.1.

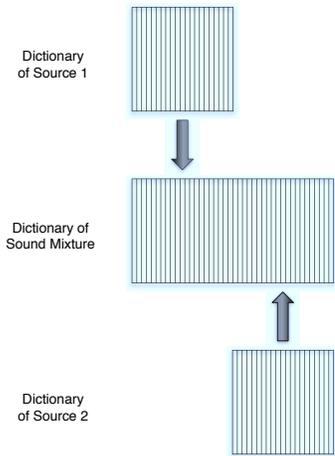

Figure 3. Combining dictionaries of two sources to model a mixture using non-negative matrix factorization.

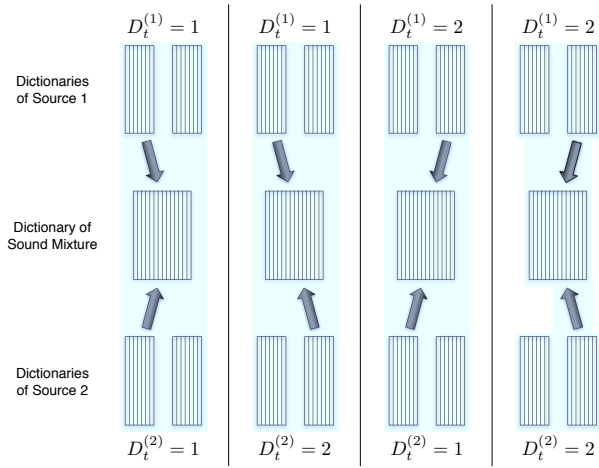

Figure 4. Combining dictionaries of two sources to model a mixture using the non-negative factorial hidden Markov model. In this simple example, each source has two dictionaries so there are a total of four ways of combining them.

NMF and its probabilistic counterparts have been used extensively for audio source separation (Virtanen, 2007; Smaragdis et al., 2007). The basic idea is to first learn a dictionary for each source from isolated training data. The mixture is then modeled by a dictionary formed from the concatenation the dictionaries of the two[1] sources (Figure 3). The goal is then to estimate mixture weights over all dictionary elements at each time frame. Using these mixture weights, we can reconstruct the contribution of each source at each time frame, obtaining the separated spectrogram of each source. The phase of the original mixture STFT is typically then used to obtain time-domain audio signals from these separated spectrograms.

The N-HMM has been extended to the non-negative factorial hidden Markov model (N-FHMM) to model sound mixtures and has been used for source separation (Mysore et al., 2010). Each time frame of the spectrogram is modeled by one of the combinatorially many concatenations of dictionaries of the two sources as illustrated in Figure 4. When used for source sep-aration, the goal is to estimate the posterior probabilities of using each pair of dictionaries at each time frame as well as the mixture weights for each of these dictionary pairs. Using these estimates, we can reconstruct the separated spectrograms and perform source separation in the same way as described above.

---

[1] It is straightforward to extend this and other methods described in the paper to more than two sources.

The N-FHMM has been shown to achieve much better source separation accuracy than simple NMF (Mysore et al., 2010). However, the combinatorial nature of the N-FHMM makes the complexity of exact inference exponential in the number of sound sources, which is often intractable. Specifically, if each source has $N$ states and there are $S$ sources, then we must evaluate the posterior probabilities of $N^S$ state configurations per time frame.

It would therefore be useful to be able to use an approximate inference technique for the N-FHMM. Structured variational inference (Ghahramani & Jordan, 1997) is an attractive approach for factorial hidden Markov models (FHMM) in general. However, the natural extension of this idea to the N-FHMM has certain limitations, which make it a poor approximation (Section 3.1). In this paper, we propose a Bayesian variant of the N-FHMM (Section 2.2) that makes it more amenable to variational inference, and then develop a suitable factored approximation to the posterior distribution deriving the corresponding variational inference algorithm (Section 3.2).

Experiments (Section 5) show that our algorithm achieves accuracy comparable to that of exact infer-ence, but is about 30 times faster on the configurations of the N-FHMM that achieve the the highest-quality source separation results.

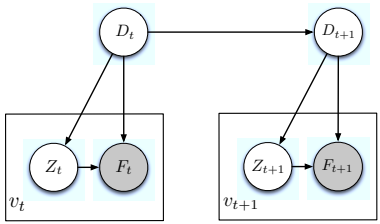

Figure 5. Graphical model of the non-negative hidden Markov model.

## 2. Probabilistic Models

In this section, we first describe the probabilistic model of the N-HMM (Mysore et al., 2010) for single sources as it forms the foundation for the N-FHMM. We then describe the probabilistic model of the proposed Bayesian variant of the N-FHMM. In these models, each time frame of the spectrogram is viewed as a histogram of "sound quanta" in the same way that a document is viewed as a histogram of words in topic models (Hofmann, 1999; Blei et al., 2003).

### 2.1. Non-negative Hidden Markov Model

The graphical model is shown in Figure 5. The random variables $D_{1...T}$ form a Markov chain, and the spectra in each time frame are independent given these variables. Each possible value of $D_t$ identifies a spectral dictionary. Each dictionary contains a set of dictionary elements (analogous to topics), one of which is selected for each sound quantum by the random variable $Z_t$. Each dictionary element is a normalized vector over frequencies (analogous to a distribution of words). The frequency associated with a particular quantum is selected by $F_t$.

The generative process at time frame $t$ is thus:

1. Choose state $D_t|D_{t-1} \sim$ Discr $[\rho(D_{t-1})]$

2. Repeat for each of $v_t$ quanta:
   – Choose dictionary element $Z_t \sim$ Discr $[\theta_t(D_t)]$
   – Choose frequency $F_t \sim$ Discr $[\beta(D_t, Z_t)]$

Here, Discr [] represents the discrete distribution; $\rho(d)$ is the column of the Markov transition matrix representing transitions from state $d$; $\theta_t(d)$ is a vector of normalized mixture weights for dictionary element $d$ in time frame $t$; and $\beta(d, z)$ is the normalized dictionary element $z$ of dictionary $d$. Given the spectrogram of a sound source, maximum–likelihood (ML) values of all these parameters may be found by the EM algorithm. The dictionaries and the transition matrix define the model of the sound source, whereas the mixture weights (which depend on $t$) are nuisance parameters unique to the particular instance of the sound source used for training. A sample of the dictionaries learned from real speech data is shown in Figure 6.

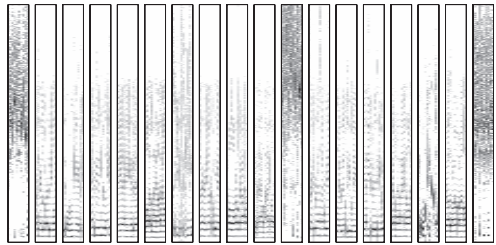

Figure 6. Non-negative hidden Markov model dictionaries that were learned from speech data. Each dictionary contains 10 dictionary elements that are stacked right next to each other. We show a subset of the 40 dictionaries that were learned in this example. We see that these dictionaries roughly correspond to subunits of speech and some are harmonic while others are noise like.

### 2.2. Non-negative Factorial Hidden Markov Model

The original N-FHMM introduced an independent Markov chain $D^{(s)}_{1...T}$ for each source $s$, and time-dependent mixing weights that selected elements from a combined state-dependent dictionary $\theta_t(d^{(1)}, d^{(2)})$. Here, we extend this model in two ways. First, $\theta_t$ and $Z_t$ will range over all dictionary elements of all sources. Thus the selection of a dictionary based on the state $D^{(s)}_t$ becomes probabilistic, and elements from more than one dictionary may appear in principle. Second, we treat $\theta_t$ as a Dirichlet-distributed latent variable, rather than as a parameter. Separating the mixture requires estimation of $\theta_t$: the older N-FHMM formulation used ML estimates; here we use a variational posterior.

The generative process (Figure 7) at time frame $t$ is thus:

1. Choose states for each source:
   $D^{(s)}_t|D^{(s)}_{t-1} \sim$ Discr $\left[\rho^{(s)}(D^{(s)}_{t-1})\right]$

2. Choose mixture weights:
   $\theta_t \sim$ Dirich $\left[\alpha(D^{(1)}_t, D^{(2)}_t)\right]$

3. Repeat for each of $v_t$ quanta:
   –Choose dictionary element: $Z_t \sim$ Discr $[\theta_t]$
   –Choose frequency: $F_t \sim$ Discr $[\beta(Z_t)]$

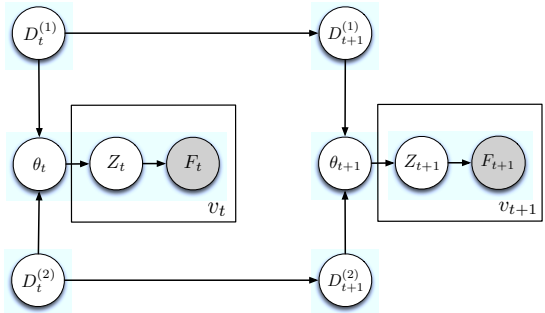

*Figure 7.* Graphical model of the proposed Bayesian variant of the non-negative factorial hidden Markov model.

The function $\alpha$ gives the Dirichlet parameters for the mixing weights, and thus specifies the dictionary elements available given each pair of source Markov states. It can most easily be written by introducing indicator variables $\delta_{t,n}^{(s)} = 1$ if $D_t^{(s)} = n$ and 0 otherwise; as well as a binary mask array $B$, with $B_{snk} = 1$ iff dictionary element $k$ is available when the Markov chain associated with source $s$ is in state $n$. Then the $k$th generative Dirichlet parameter for $\theta_t$ can be written $\alpha_k(D_t^{(1)}, D_t^{(2)}) = 1 + \gamma \sum_s \sum_n \delta_{t,n}^{(s)} B_{snk}$. Provided the two sources do not share dictionary elements, the sum in this expression evaluates to either 0 or 1. Thus the distribution on $\theta_t$ has parameters of $1 + \gamma$ corresponding to the elements selected by the current Markov states, and 1 otherwise. The hyperparameter $\gamma$ sets the concentration of the Dirichlet; we took $\gamma = 1$. Thus, we can write the distribution on mixing weights:

$$P(\theta_t | D_t^{(1)}, D_t^{(2)}) \propto \prod_k \left( \prod_s \prod_n \theta_{t,k}^{\delta_{t,n}^{(s)} B_{snk}} \right). \quad (1)$$

This formulation does not make exact inference any easier because the number of such distributions that we have to consider is still exponential in the number of sources. However, we will see in Section 3.2 that this is important for our variational inference derivation.

The remaining parameters are much as before: $\beta(z)$ is the $z$th normalized dictionary element; $\rho^{(1)}$ and $\rho^{(2)}$ are the transition matrices of the two sources. We also need parameters for the initial state probabilities, for which we write $\pi^{(1)}$ and $\pi^{(2)}$. The number of quanta at each time frame $v_t$ could be modeled as a draw from (say) a Poisson distribution. However, it is independent of the other generative variables and (in our applications) is observed, so we do not model it as a random variable.

Without the temporal dynamics, our formulation is similar to that of latent Dirichlet allocation LDA (Blei et al., 2003). The main difference is that the Dirichlet distribution in a given time frame is a function of the Markov states of the sources rather than being constant for all time frames.

## 3. Variational Inference

The parameters describing each source N-HMM are learned from isolated training data of that source. Thus, the goal of inference in the N-FHMM is only to resolve the mixture; specifically, to estimate the marginalized posterior distribution of the mixture weights $P(\theta_t | \bar{\mathbf{f}})$ at each time frame. Once this distribution is found, we can reconstruct the individual sources and therefore perform source separation. The full posterior distribution is given by $P(\bar{\mathbf{Z}}, \bar{\theta}, \overline{D^{(1)}}, \overline{D^{(2)}} | \bar{\mathbf{f}})$, where $\bar{\theta}$, $\overline{D^{(1)}}$, and $\overline{D^{(2)}}$ represent $\theta_t$, $D_t^{(1)}$, and $D_t^{(2)}$ at all time frames and $\bar{\mathbf{Z}}$ represents all draws of $Z_t$ at all time frames. $\bar{\mathbf{f}}$ represents the observed values of $F_t$ at all time frames. The computational cost of finding the posterior distribution is exponential in the number of sources due to the coupling of the states of the individual N-HMMs. Exact inference in N-FHMMs is therefore intractable so we resort to approximation.

Variational inference (Jordan et al., 1999) refers to a class of techniques that are used to approximate an intractable posterior distribution with a simpler (typically factorized) distribution. By minimizing the KL divergence between the two distributions, a lower bound on the log-likelihood is maximized. This is the class of approximations that we employ.

A natural variational approximation to the N-FHMM would be to decouple from each other the sets latent variables that correspond to each component N-HMM, but to retain the structured posterior over each separate source. This scheme is analogous to the structured variational approximation for FHMMs (Ghahramani & Jordan, 1997). Unfortunately, however, it performs poorly for the N-FHMM of (Mysore et al., 2010). We first briefly sketch the approach and explain why it seems to fail, before moving to the new variant of the N-FHMM to derive a more successful variational inference algorithm.

### 3.1. Difficulties with Decoupling

Decoupling the variational posteriors for each sound source requires that it be possible to group latent variables according to the generative source. This is easy for $\overline{D^{(s)}}$. However, the latent variables $\bar{\mathbf{Z}}$ identify elements from a combined dictionary over both sources. Thus, to proceed we introduce a new latent variable $S_t$ to indicate the proportions of quanta drawn from each source at time $t$, and then separately generate $\overline{\mathbf{Z}^{(1)}}$ and $\overline{\mathbf{Z}^{(2)}}$, each ranging over the dictionary of a single source. In this parameterization, the posterior is $P(\overline{\mathbf{Z}^{(1)}}, \overline{D^{(1)}}, \overline{\mathbf{Z}^{(2)}}, \overline{D^{(2)}}, \overline{S}|\overline{\mathbf{f}})$, which might be approximated by the decoupled variational distribution:

$$q(\overline{\mathbf{Z}^{(1)}}, \overline{D^{(1)}})q(\overline{\mathbf{Z}^{(2)}}, \overline{D^{(2)}}) \prod_t q(S_t).$$

In this form, the two components $q(\overline{\mathbf{Z}^{(1)}}, \overline{D^{(1)}})$ and $q(\overline{\mathbf{Z}^{(2)}}, \overline{D^{(2)}})$ do indeed correspond to structured posteriors within the two N-HMMs describing the sound sources, while the factor $q(S_t)$ corresponds to the mixing proportions of the two sources at time frame $t$. The variational iterations then update each individual N-HMM posterior using the forward–backward algorithm (Rabiner, 1989) while keeping the contribution of the other N-HMM fixed; and then revise the mixing proportions of the sources. While certainly plausible, this algorithm proves to be very prone to sticking in local optima and in experiments performs more poorly than even basic NMF (implemented in a probabilistic form, see Section 5). Here we provide an intuitive sketch of what we see as the source of the difficulty.

In many applications, the sound sources to be separated may have some spectrally similar aspects, so that some or all of the dictionary elements in their individual N-HMMs may have similar forms. This is the case when the sound sources are, for example, speech from different speakers. In such a situation, the dictionary elements of one source may be able to provide a reasonable fit to sounds generated by the other source.

Consider an example of a single time frame in a speech mixture in which the first source contributed a harmonic spectrum (say a vowel) while the second source produced a noise-like spectrum as might be associated with a fricative. Both source models are likely to contain spectral dictionary elements to account for both vowels and fricatives, although these elements might belong to different dictionaries within each source model. Thus, if at an early stage the harmonic structure is incorrectly assigned to source 2 and the noise-like component to source 1, the inferred Markov states for the two components will be incorrect. We find that it is then very unlikely that further iterations will resolve the error, indeed they seem to make it worse. As the incorrect assignments reinforce each other in the two models, the posterior over Markov states becomes very sharp. Thus, the two sources are confused. This reflects a local optimum: it could very well be that the variational free energy would be larger for the correct assignment, but the hill-climbing form of the iterative algorithm makes it unable to discover that fact.

In experiments, we found that this situation appeared with some frequency, despite the fact that the temporal structure of the underlying Markov process biased solutions away from such confusions to an extent.

### 3.2. Proposed Variational Approximation

In the proposed variant of the N-FHMM, the link between Markov state and dictionary element is less absolute. Also, we estimate a full posterior over the mixing proportions $\theta_t$ over all dictionary elements, rather than obtaining an ML point estimate—this reduces the risk of zeros (or very small values) in the point estimate, which would have created a similar barrier to exploration. Thus both sources are able to explore the full range of possible dictionary elements and settle on the correct apportionment of the mixture spectrum, while the interaction between Markov state variables and prior on $\theta_t$ strongly favors explanations that concentrate on a single dictionary per source.

To develop the variational algorithm for this model, we approximate the posterior distribution $P(\overline{\mathbf{Z}}, \overline{\theta}, \overline{D^{(1)}, D^{(2)}}|\overline{\mathbf{f}})$ with the following factored form:

$$\left(\prod_t q(\theta_t)\right) \left(\prod_t \prod_v q(Z_{t,v})\right) q(\overline{D^{(1)}})q(\overline{D^{(2)}}).$$

Note that for a given time frame, the index $k$ for $\theta_t$ is over all dictionary elements of all dictionaries of all sources. The mixture weights for a given time frame $t$ are therefore in a single factor $q(\theta_t)$. Moreover, this factor is independent of $D_t^{(1)}$ and $D_t^{(2)}$ so we do not have the combinatorial problem.

However, the distribution over the states of all time frames of a given sound source $q(\overline{D^{(1)}})$ and $q(\overline{D^{(2)}})$ are each a single factor. This is because we would still like to make use of the structure of the temporal dynamics in each individual source and exact inference is efficient using the forward–backward algorithm.

By minimizing the KL divergence between the true posterior distribution and the factorized distribution, we obtain the following variational inference solution (Jordan et al., 1999) for each of the factors:

$$q(\theta_t) \propto \exp \left\langle \log P(\overline{\mathbf{Z}}, \overline{\theta}, \overline{D^{(1)}, D^{(2)}}, \overline{\mathbf{f}}) \right\rangle_{q_1}, \quad (2)$$

$$q(Z_{t,v}) \propto \exp \left\langle \log P(\overline{\mathbf{Z}}, \overline{\theta}, \overline{D^{(1)}, D^{(2)}}, \overline{\mathbf{f}}) \right\rangle_{q_2}, \quad (3)$$

$$q(\overline{D^{(1)}}) \propto \exp \left\langle \log P(\overline{\mathbf{Z}}, \overline{\theta}, \overline{D^{(1)}, D^{(2)}}, \overline{\mathbf{f}}) \right\rangle_{q_3}, \quad (4)$$

$$q(\overline{D^{(2)}}) \propto \exp \left\langle \log P(\overline{\mathbf{Z}}, \overline{\theta}, \overline{D^{(1)}, D^{(2)}}, \overline{\mathbf{f}}) \right\rangle_{q_4}, \quad (5)$$

where $q_1$ refers to the product of all of the factors except $q(\theta_t)$ (and similarly for the other factors). We use proportionality rather than equality to simply indicate that the quantities are unnormalized. Solving these equations, we find that $q(\theta_t)$ is a Dirichlet distribution, which we parameterize by $\widehat{\alpha}_{t,k}$; $q(\overline{D^{(1)}})$ and $q(\overline{D^{(2)}})$ are each a set of discrete distributions, with marginal probabilities $\widehat{d}_{t,n}^{(1)}$ and $\widehat{d}_{t,n}^{(2)}$ at time frame $t$. The distribution $q(Z_{t,v})$ is also discrete, and at any time frame the parameters for all $Z_{t,v}$ whose corresponding observed frequencies $f_{t,v}$ are equal, will be identical. Thus, we write these parameters as $\widehat{z}_{t,l,k}$, where $l$ indexes frequency (in place of $v$), and $k$ identifies a dictionary element.

On simplification of Eq.2, we obtain following update equation for the parameters of $q(\theta_t)$:

$$\widehat{\alpha}_{t,k} = \sum_v \widehat{z}_{t,f_{t,v},k} + \gamma \sum_s \sum_n \widehat{d}_{t,n}^{(s)} B_{snk} + 1 \,.$$

Note that as $\widehat{d}_{t,n}^{(s)}$ are defined to be marginal probabilities, they are exactly the expected values under $q$ of $\delta_{t,n}^{(s)}$. The index $f_{t,v}$ is the observed frequency of quantum $v$ at time $t$. We can group all quanta for which $f_{t,v}$ is equal, to obtain:

$$\widehat{\alpha}_{t,k} = \sum_l V_{lt} \widehat{z}_{t,l,k} + \gamma \sum_s \sum_n \widehat{d}_{t,n}^{(s)} B_{snk} + 1 \,, \quad (6)$$

where $V_{lt}$ is the value of the spectrogram (i.e., the number of quanta) at frequency $l$ and time frame $t$.

On simplification of Eq.3, we obtain estimates of the parameters of $q(Z_{t,v})$:

$$\log \widehat{z}_{t,l,k} = \log \beta_l(k) + \psi(\widehat{\alpha}_{t,k}) - \psi\Big(\sum_k \widehat{\alpha}_{t,k}\Big) - \kappa \,, \quad (7)$$

where $\psi()$ is the digamma function; $\kappa$ is a log normalizer; and $\beta_l(k)$ is the value of dictionary element $k$ at frequency $l$. The digamma terms arise from the normalizing $\Gamma$-functions of the Dirichlet distribution (Blei et al., 2003).

On simplification of Eq.4 and 5, we first obtain surrogate "likelihood" terms, which we subsequently use in the forward–backward algorithm to obtain the distribution parameters. This likelihood term at time frame $t$ for state $n$ of source $s$ is given by:

$$\widehat{\phi}_{t,n}^{(s)} = \sum_k B_{snk} \left( \psi(\widehat{\alpha}_{t,k}) - \psi\Big(\sum_k \widehat{\alpha}_{t,k}\Big) \right) \,. \quad (8)$$

The forward–backward algorithm then finds estimates of the marginals $\widehat{d}_{t,n}^{(s)}$ of $q(\overline{D^{(s)}})$.

We iterate over Eqs.6,7,8, and the forward–backward algorithm for each source. The resulting solution provides estimates of the parameters $\widehat{\alpha}_{t,k}$ of the distribution $q(\theta_t)$, indicating the distribution of mixture weights.

## 4. Source Separation

We reconstruct the spectrograms of the individual sources by taking linear combinations of the dictionary elements of all dictionaries of each individual source according to the estimated mixture weights $\widehat{\alpha}_{t,k}$, at each time frame. This gives us estimates of the separated spectrograms of each source $\widehat{V}_{lt}^{(s)}$. We can simply go back to the time domain with these estimates using the phase of the original mixture. However, a common source separation practice to first obtain more refined spectrogram estimates by applying the following masking strategy:

$$V_{lt}^{*(s)} = V_{lt} \frac{\widehat{V}_{lt}^{(s)}}{\sum_{s'} \widehat{V}_{lt}^{(s')}},$$

where $V_{lt}$ is the original mixture spectrogram. The final estimated spectrogram for each source is therefore $V_{lt}^{*(s)}$. We employ this strategy in our experiments.

## 5. Experimental Results

To validate our proposed variational inference algorithm, we performed source separation experiments on speech mixtures and compared our results to those of exact inference within the N-FHMM of (Mysore et al., 2010). As the proposed algorithm has much lower computational complexity, our goal was to try to achieve source separation performance that came close to that of exact inference. Additionally, we compared to the performance of the decoupled variational approximation (Section 3.1), and to the performance of probabilistic latent component analysis (PLCA) (Smaragdis et al., 2007), this being the baseline comparison used in (Mysore et al., 2010). PLCA is a probabilistic audio interpretation of NMF (up to a column-wise normalization). It therefore serves as an baseline equivalent to NMF.

We performed experiments with 50 different speech mixtures and report the mean of the results. Data were taken from TIMIT, a commonly used corpus for speech processing and speech recognition tasks. It comprises numerous sentences from multiple speakers.

We performed each of the 50 experiments as follows. We first randomly chose one male and one female speaker. For each speaker, we assigned nine sentences as training data and one sentence as test data. We then concatenated the training data sentences and ob-tained a spectrogram with a window size of 64ms and a hop size of 16ms (the sampling rate was 16KHz). Next, we learned an N-HMM (with an ergodic Markov chain) from the training data of each speaker, yielding a set of dictionaries and a transition matrix for each speaker. The next step was to mix the two test sentences (one from each speaker) at 0dB and obtain the spectrogram using the above window size and hop size. We then combined the dictionaries and transition matrices of the two speakers into either a joint PLCA or an N-FHMM model, performed inference using the various methods, and separated the sources.

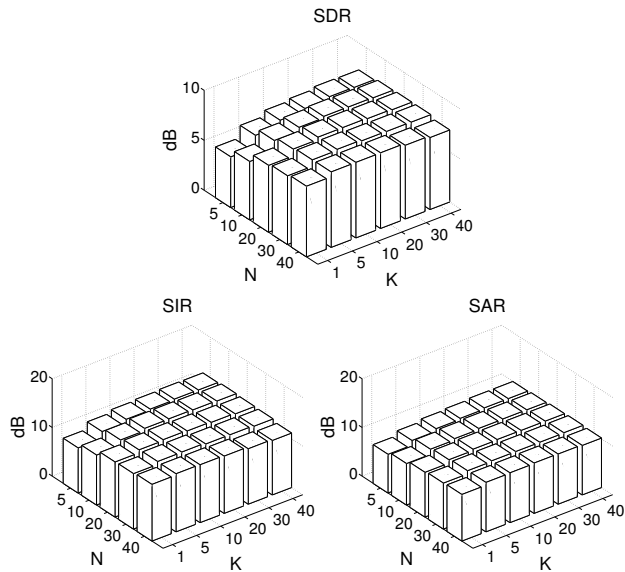

Figure 8. Source separation metrics for different configurations of the proposed method. N represents the number of dictionaries and K represents the number of elements per dictionary.

|  | SDR | SIR | SAR |
|---|---|---|---|
| PLCA | 5.73 | 9.53 | 8.78 |
| Exact | 7.73 | 14.05 | 9.27 |
| Decoupled Variational | 2.10 | 3.40 | 10.10 |
| Proposed Variational | 7.56 | 11.87 | 10.22 |
| Proposed Variational Opt. | 7.63 | 12.00 | 10.24 |

Table 1. BSS-EVAL source separation performance for various algorithms. PLCA used a dictionary of 30 elements. N-FHMM with exact inference, decoupled variational inference, and the proposed variational inference algorithm all used 20 dictionaries of 30 elements each. The final line shows the proposed method with 20 dictionaries of 20 elements each, which was the optimal configuration for that method.

We used the standard BSS-EVAL suite of metrics (Vincent et al., 2006) to evaluate the source separation performance. This suite consists of three signal to noise ratio (SNR) type metrics (in dB). Source to Interference Ratio (SIR) evaluates the suppression of the unwanted source. Source to Artifacts Ratio (SAR) evaluates the amount of artifacts introduced by the separation process (with larger numbers reflecting less artifacts). Source to Distortion Ratio (SDR) gives us an overall source separation score that takes both the suppression capability as well as the introduced artifacts into account. We computed these metrics on each of our 100 sources separated from 50 mixtures.

In order to find the optimal configuration of our model (the number of dictionaries and number of elements per dictionary), we repeated these experiments in 30 different configurations. The BSS-EVAL metrics for all of these configurations are shown in Figure 8. The optimal configuration (in terms of SDR) was 20 dictionaries with 20 elements each. It is evident, however, that the different configurations yield similar performance scores, except for a noticeable drop when using only 1 element per dictionary. This is encouraging as it implies that the algorithm is not particularly sensitive to the specific configuration for this kind of data.

To make comparison to exact inference unbiased, we also searched for the optimal configuration in that case. Here, 20 dictionaries with 30 elements each yielded the best source separation performance (data not shown), although, as with the proposed variational inference algorithm, the metrics did not vary substantially with different configurations. Table 1 shows the results obtained with this configuration as well as the results of the proposed method and decoupled variational inference when using the same configuration.

As a baseline, we also experimented with PLCA with various dictionary sizes, and found that 30 dictionary elements yielded the optimal source separation performance. The results of using this configuration of PLCA are are also shown in Table 1.

We see that our algorithm performs almost as well as exact inference even when using the same configuration (Table 1). The difference in SDR is less than 0.2 dB. There is however a large difference in computation time. Based on the configuration of each source having 20 dictionaries of 30 elements each, we empirically found each iteration of the proposed method to be about 30 times faster than using exact inference. However, when using the optimal configuration for the proposed method (20 dictionaries of 20 elements each), we observed about a 40x speedup. The proposed method generally took about twenty iterations

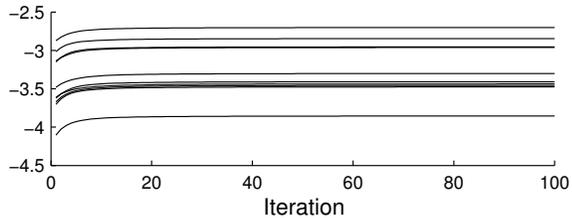

Figure 9. Reconstruction cross entropy at each iteration of the proposed variational inference algorithm for ten different source separation experiments. There is usually convergence within the first twenty iterations.

to converge (Figure 9), which is a similar number to that seen with exact inference.

The SIR of the proposed method is lower than that of exact inference but the SAR is higher. This can be understood as follows. Exact inference returns a higher SIR because it is more constrained. Only one dictionary from each source may be active in each time frame. This restriction is relaxed in the proposed method, allowing some interference from elements of the other dictionaries. This very property gives the proposed method a higher SAR. In order to reduce artifacts, it can be helpful to recruit some contribution from dictionary elements that correspond to non-active Markov states. This can help explain nuances in the spectral time frame that the active dictionary might not capture completely. This is possible in the proposed method, but not in the rigid exact N-FHMM model. The proposed method and exact inference therefore have a fairly even SIR/SAR trade off leading to approximately the same SDR scores.

As shown in Table 1, the proposed method outperforms PLCA in all three metrics and the decoupled variational approximation performs very poorly with a lower SDR and SIR than even PLCA.

## 6. Conclusions

We have proposed a Bayesian variant of the N-FHMM and an efficient variational inference algorithm for the model. The computational complexity of the algorithm is linear in the number of sources, and it is about 30 times faster than exact inference on an empirically optimal configuration of the N-FHMM, with comparable source separation accuracy. Although variational inference in the N-FHMM was demonstrated on the task of source separation, it is a general model of sound mixtures and can be used for various other audio tasks such as concurrent speech recognition of multiple speakers and automatic music transcription.